\documentclass[runningheads]{llncs}
\usepackage[T1]{fontenc}
\usepackage{graphicx}
\usepackage{cite}
\usepackage{amsmath,amssymb,amsfonts}
\usepackage{algorithmic}
\usepackage{multirow}
\usepackage{textcomp}
\usepackage{booktabs}
\usepackage{xcolor}
\usepackage{float}

\begin{document}
%


%
\title{CrossGP: Cross-Day Glucose Prediction Excluding Physiological Information}
\titlerunning{CrossGP: Cross-Day Glucose Prediction Excluding Physiological Information}
\author{Ziyi Zhou$^{*\dag1}$ \and Ming Cheng$^{*1}$ \and Yanjun Cui$^{1}$ \and Xingjian Diao$^{1}$ \and Zhaorui Ma$^{2}$}
\authorrunning{Zhou et al.}

\institute{
$^{1}$ Dartmouth College, 15 Thayer Drive, Hanover, NH 03755, USA\\
\email{ziyi.zhou.gr@dartmouth.edu}\\
$^{2}$ George Mason University, 4400 University Drive, Fairfax, VA 22030, USA\\
\email{zma4@gmu.edu}
}

\maketitle

\begin{abstract}
The increasing number of diabetic patients is a serious issue in society today, which has significant negative impacts on people's health and the country's financial expenditures.
Because diabetes may develop into potential serious complications, early glucose prediction for diabetic patients is necessary for timely medical treatment. 
Existing glucose prediction methods typically utilize patients' private data (e.g. age, gender, ethnicity) and physiological parameters (e.g. blood pressure, heart rate) as reference features for glucose prediction, which inevitably leads to privacy protection concerns. 
Moreover, these models generally focus on either long-term (monthly-based) or short-term (minute-based) predictions. Long-term prediction methods are generally inaccurate because of the external uncertainties that can greatly affect the glucose values, while short-term ones fail to provide timely medical guidance. 
Based on the above issues, we propose CrossGP, a novel machine-learning framework for cross-day glucose prediction solely based on the patient's external activities without involving any physiological parameters. 
Meanwhile, we implement three baseline models for comparison. Extensive experiments on Anderson's dataset strongly demonstrate the superior performance of CrossGP and prove its potential for future real-life applications.

\keywords{cross-day glucose prediction \and physiological parameter \and machine learning \and privacy protection.}
\end{abstract}

\newcommand\blfootnote[1]{%
  \begingroup
  \renewcommand\thefootnote{}\footnote{#1}%
  \addtocounter{footnote}{-1}%
  \endgroup
}

\blfootnote{$^*$Equal contribution. $^\dag$Corresponding author.}

\section{Introduction}

Recently, around $25.5$ million people in the U.S. have been diagnosed with diabetes, representing $7.6\%$ of the total U.S. population \cite{parker2024economic}.
This number is expected to increase, reaching $47.1$ million people in the U.S. and $578$ million people worldwide by 2030 \cite{saeedi2019global}.
Meanwhile, 
the total estimated cost of diagnosed diabetes in the U.S.
are enormous, with $412.9$ billion U.S. dollars spent \cite{parker2024economic}. 
Therefore, the current situation of the diabetic epidemic is extremely serious, and controlling blood glucose levels in diabetic patients is not only essential for their health, but also for the country's financial expenditures.
Accurate glucose prediction is essential for offering informative and timely clinical guidance to diabetic patients, providing meaningful guidance in various application scenarios including health monitoring \cite{li2022smart, sachmechi2023frequent, wang2024differential}, wearables development \cite{saha2023wearable, AlikhaniKoshkak2024SEAL}, and smart healthcare \cite{wang2020guardhealth, zhou2021doseguide, zhang2023doseformer, li2021fully}. 

Existing methods of glucose prediction are typically based on the combination of the patient's physiologic indicators \cite{butt2023feature, turksoy2013hypoglycemia} (e.g. heart rate, skin temperature, GSR) and anthropogenic activities \cite{munoz2020deep, li2019convolutional} (e.g. sleep, meal intake, insulin doses) to predict future glucose values. 
There are also methods \cite{aldhoon2014glucose, stojmenski2023age, bergenstal2017racial} utilizing more private personal information, including gender, age, race, and BMI, for accurate blood glucose prediction. 
However, although these methods integrate multiple features for accurate blood glucose prediction, \textit{\textbf{they have to collect patients' private data}}, potentially raising ethical issues along with privacy protection concerns \cite{de2023federated, piao2023blood, cheng2023saic}. 
In addition, obtaining patient information requires sophisticated measuring instruments, as well as considerable human resources \cite{so2012recent}. 

Moreover, blood glucose prediction models at the current stage can be categorized into \textit{\textbf{two}} \textit{\textbf{groups}}: long-term prediction and short-term prediction.
Long-term prediction methods \cite{liu2019long, jaloli2023long} generally predict the patient's blood glucose values for the next month or even beyond, mainly focusing on the prediction of the general trend. Therefore, the prediction of specific glucose values is relatively inaccurate, as numerous uncertain factors may affect the patient's blood glucose level over a long time interval.
In contrast, short-term prediction methods \cite{dassau2010real, ben2015identification, munoz2020deep} specialize in predicting glucose values over a short period (e.g. $10-30$ minutes). Although high prediction accuracy can be achieved by combining the patient's physiologic parameters, the results are not significantly informative for clinical guidance because the prediction interval is too short for effective interventions and therapeutic measures.
Therefore, there remains an unsolved problem: How to accurately predict blood glucose values for an appropriate interval (e.g. one day) without utilizing the private data of diabetic patients to provide effective clinical guidance?

Addressing these challenges is essential for both diabetes management and broader medical research~\cite{wei2023end, ng2022predicting, pedram2023experience}. Benefiting from the powerful ability of machine learning and neural networks to learn high-dimensional features for accurate prediction, we propose CrossGP -- a novel framework for diabetic patients' cross-day glucose prediction without involving personal anthropometric and biochemical data. In summary, our contribution is threefold:

\begin{figure*}[htbp]
  \centering
  \includegraphics[width=0.9\textwidth]{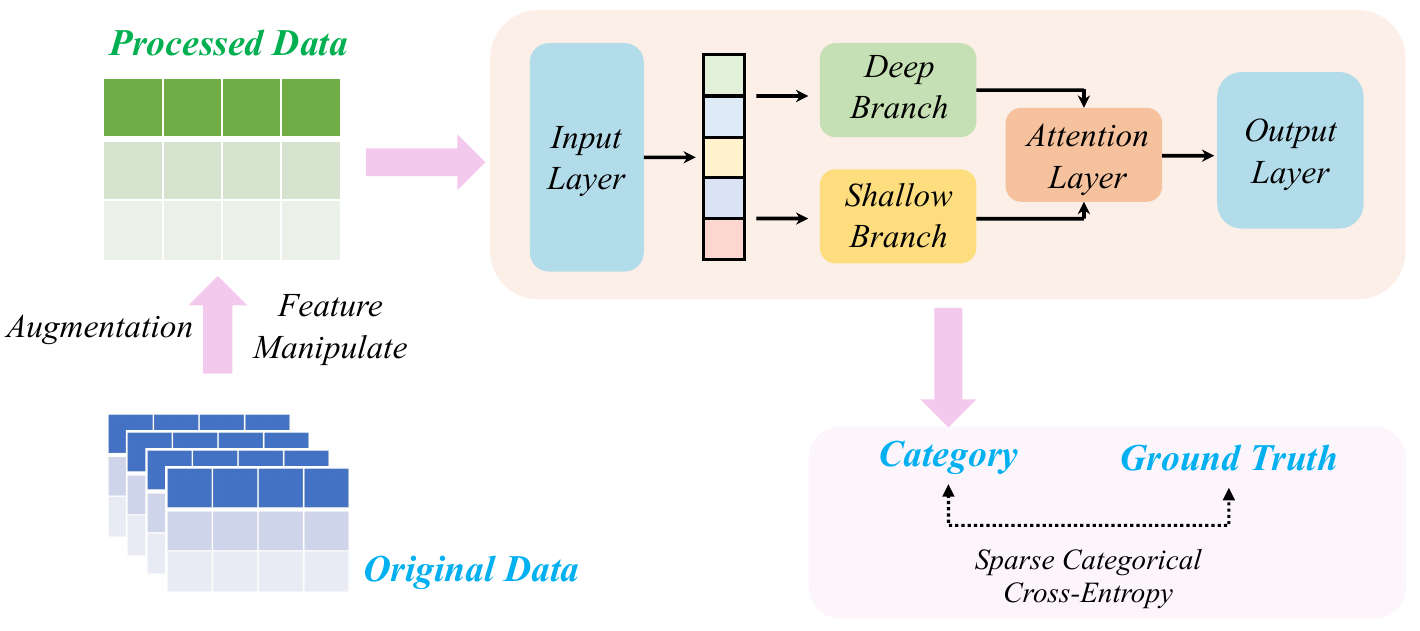}
  \caption{\textbf{
  Overview architecture of CrossGP.} 
  (1) We pre-process the original data (e.g. records merging) and employ data augmentation to enhance the model's robustness. 
  (2) The processed data is input into the model to predict the corresponding class. The deep/shallow branch specializes in capturing features in multi-scale, followed by an attention layer to fuse the feature. 
  (3) Following the standard criterion, the cross-entropy loss is constructed between predicted categories and the ground truth labels. 
  }
  \label{fig:arch}
\end{figure*}

\begin{enumerate}
    \item {
    \textbf{A novel and effective framework for cross-day glucose prediction.} 
    To the best of our knowledge, CrossGP is the first framework for cross-day glucose prediction on Anderson's dataset \cite{anderson2016multinational}, showcasing its novelty and innovation.  
    }
    \item {
    \textbf{Effective glucose prediction results. }
    We implement several comparable machine learning baselines to evaluate the performance, where CrossGP outperforms all baselines and reaches cutting-edge performance. 
    }
    \item {
    \textbf{Essential privacy protection. }
    No private information (e.g. age, gender, race) or physiologic data (e.g. heart rate, GSR, blood pressure) is involved in CrossGP for model training, protecting patient privacy while realizing high-precision blood glucose prediction. 
    }
\end{enumerate}

\section{Methods}
\subsection{Logistic Regression}
As a commonly used statistical algorithm for prediction tasks, Logistic Regression (LR) \cite{lavalley2008logistic} models an event using a linear relationship between input features and output. Formally, given $X = \{x_1, x_2, ..., x_n\}$ as the feature input, the model outputs the probability within the range of $(0, 1)$ representing the probability of the feature falling into a particular category. 
\begin{equation}
    \mathcal{F}(X) = \frac{1}{1 + e^{-(\beta_0 + \beta_1X)}}
\end{equation}
where $\beta_0$ is the intercept while $\beta_1$ indicates the coefficient vectors.
The parameters of LR are estimated using maximum likelihood estimation (MLE). This involves fitting the model to training data in a manner that maximizes the likelihood of observing the provided outcomes, given the assumed logistic distribution.

\subsection{Random Forest}
As an ensemble learning method widely used for prediction and classification, Random Forest (RF) \cite{breiman2001random} operates by constructing a multitude of decision trees during training. Each tree is constructed independently and makes its prediction, and the final prediction is determined by aggregating the predictions of all the trees. Formally, given $\hat{Y} = \{\hat{Y}_1, \hat{Y}_2, ..., \hat{Y}_N\}$ as the predictions generated by $N$ decision trees, the final output of RF ($\hat{Y}_{RF}$) is defined as: 
\begin{equation}
    \hat{Y}_{RF} = \frac{1}{N}\sum_{i=1}^{N}\hat{Y}_i
\end{equation}

\subsection{XGBoost}
XGBoost (Extreme Gradient Boosting) \cite{chen2016xgboost} is based on the gradient boosting framework, which builds an ensemble of weak learners (decision trees) sequentially. XGBoost corrects errors made by previous decision trees to lead to a more accurate and robust final model.
Specifically, the model assigns weights to decision trees, with subsequent decision trees having increased weight values if the current one predicts incorrectly. Afterward, these individual trees ensemble to form a more precise model. 

The objective function of XGBoost is defined as follows:
\begin{equation}
    \mathcal{F}(\theta) = \sum_{i=1}^N[l(y_i, \hat{y}_i) + \sum_{k=1}^K\Omega(f_k)] 
\end{equation}
where $y_i$ and $\hat{y}_i$ indicate the ground truth and predicted instance, respectively, $l(\cdot)$ is the loss function, $f_k$ represents the $k^{th}$ tree in the ensemble, and $\Omega(f_k)$ refers to the regularization term penalizing the $k^{th}$ tree. The objective function $\mathcal{F}(\theta)$ is optimized during the training process.

\subsection{CrossGP (Ours)}
The overall architecture of the CrossGP framework is shown in Figure \ref{fig:arch}.
\subsubsection{Data Processing}
Since the original Anderson's dataset \cite{anderson2016multinational} disperses different features (e.g. meal intake, insulin dose) in different sub-data, we first apply feature manipulation. 
Specifically, because the data points are recorded with high frequency (e.g. CGM data is recorded every 5 minutes), we merge all entries within the same day together and create a new one. Details will be introduced in Section \ref{sec:merge}.
To enhance the model's robustness and improve the prediction accuracy, we apply data augmentation by adding noise $n \sim \mathcal{N}(0, \delta^2)$ to the data. The augmented data are input into the model for the glucose prediction task. 

\subsubsection{Model Architecture}
As shown in Figure \ref{fig:arch}, the model contains an input layer, a dual-branch structure, and an output layer. Specifically, the lightweight input layer is designed with a single-layer \textit{Dense} module for initial feature extraction. 
Inspired by \cite{hu2020dasgil}, a multi-scale feature extractor is constructed. 
In particular, a deep branch and shallow branch are employed for capturing features in multi-scale. Both deep and shallow branches are designed with sequential \textit{Dense} layers (but different depths) connected with \textit{BatchNorm} \cite{ioffe2015batch} for stabilizing learning. The deep branch (4-layer \textit{Dense}) focuses on extracting deep features with hierarchical patterns and high-level abstraction, while the shallow branch (2-layer \textit{Dense}) specializes in low-level and basic representations. 
As discussed in previous work \cite{bahdanau2014neural, diao2023av, wu2021fastformer, diao2023ft2tf}, the attention module enables the model to selectively focus on the main components that are most relevant to the task. Therefore, the multi-scale features are then fused through the attention module, followed by a lightweight output layer (single-layer \textit{Dense}) to get predictions. 
Following the standard criterion \cite{zhang2018generalized, mao2023cross}, we employ Cross-Entropy loss between the predicted class and the ground truth to guide the model training.

\section{Experiments}
\subsection{Dataset Description}

\begin{figure}[tb]
  \centering
  \includegraphics[width=0.95\linewidth]{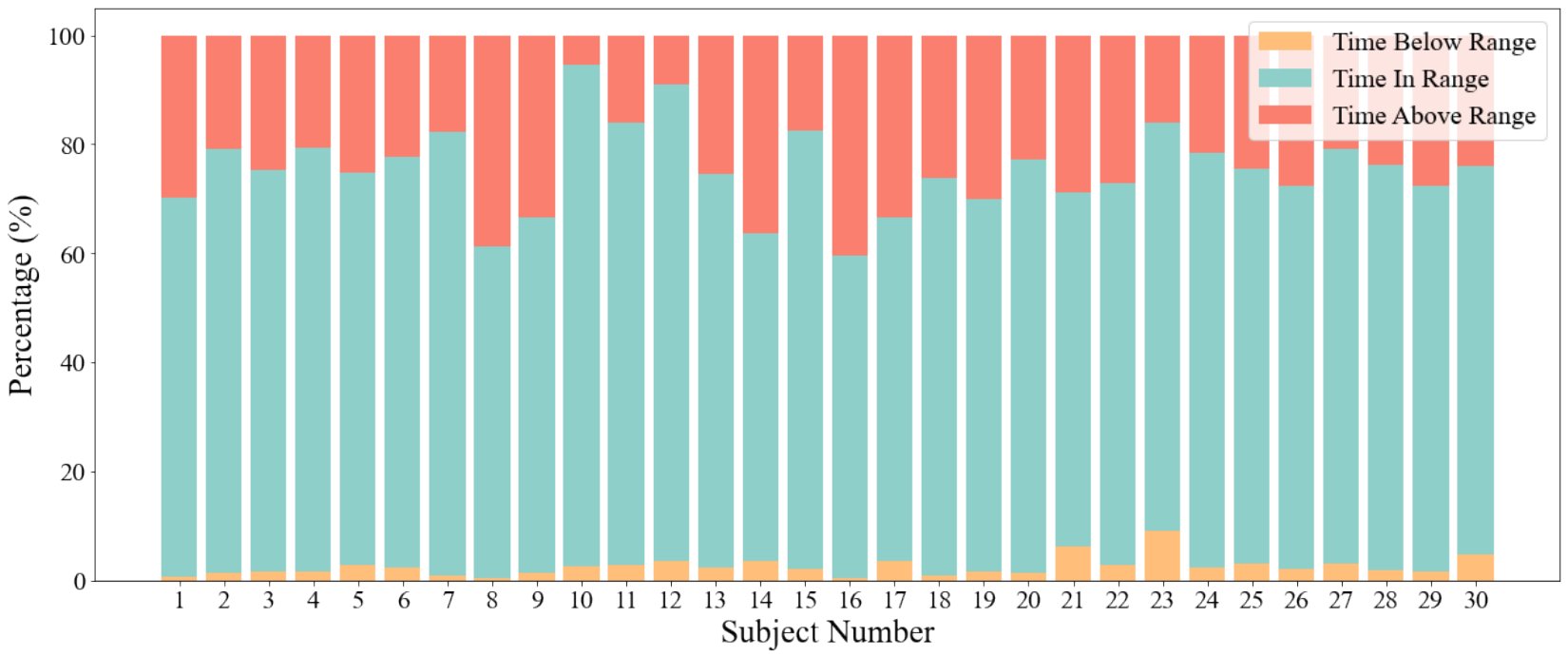}
  \caption{\textbf{
  Visualization of the CGM data.
  } 
  The percentage of TIR, TBR, and TAR of 30 subjects is shown, where TIR and TAR predominate in the figure.
  }
  \label{fig:cgm_overview}
\end{figure}

\begin{figure*}[tb]
  \centering
  \includegraphics[width=\linewidth]{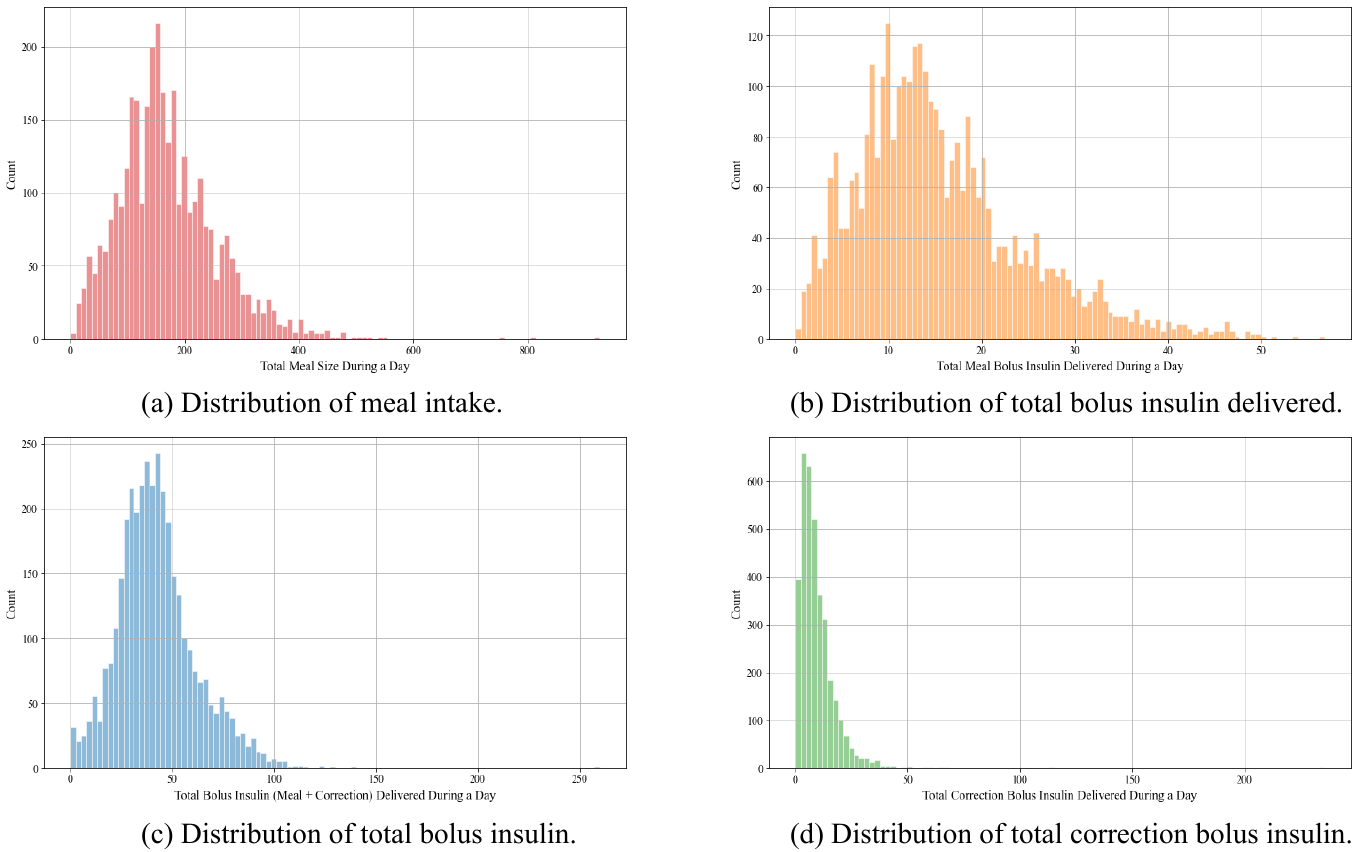}
  \caption{\textbf{
  Visualization of the feature vectors.
  } 
  The distribution of four features (meal, meal bolus, correction bolus, total bolus) of 30 subjects are demonstrated. 
  }
  \label{fig:distribution}
\end{figure*}

We thoroughly investigate Anderson's dataset \cite{anderson2016multinational} for the experiments, as summarized in Table \ref{table:summary}. This dataset is originally designed to test a closed-loop control-to-range (CTR) artificial pancreas (AP) system, containing data points of $30$ patients with Type 1 Diabetes from 03/2012 to 08/2015.
It records multiple anthropometric parameters of the patients, including weight, age, gender, ethnicity, ketone level, etc. However, since our model is designed not to involve any physiological privacy information, we only use the features listed in Table \ref{table:summary} as the information source:
\begin{itemize}
    \item {
    \textbf{CGM:} Continuous glucose monitor during a control cycle and recorded on the DiAs device.
    }
    \item {
    \textbf{Correction Bolus:} Correction bolus insulin delivered during a control cycle and recorded on the DiAs device.
    }
    \item {
    \textbf{Meal:} Meal-related data announced to the DiAs by the user and recorded on the DiAs device.
    }
    \item {
    \textbf{Meal Bolus:} Meal bolus insulin delivered during a control cycle and recorded on the DiAs device.
    }
    \item {
    \textbf{Total Bolus:} Total bolus insulin (meal, correction) delivered during a control cycle and recorded on DiAs device.
    }
\end{itemize}

The distribution of the features above are demonstrated in Figure \ref{fig:distribution}, where they span a broad range of values, representing different physiological conditions of the patients, which illustrates the dataset's diversity and robustness. 

\begin{table}[htbp]
\centering
\caption{Summary of The Original Dataset.}
\label{table:summary}
\begin{tabular}{@{}cccccc@{}}
\toprule
\textbf{Features} & \textbf{\# Data} & \textbf{Freq} & \textbf{Min} & \textbf{Max} & \textbf{Mean$\pm$STD} \\ \midrule
CGM                  & 915,827       & 5 min          & 39.0           & 401.0          &       149.4 $\pm$ 52.5              \\
Correction Bolus     & 21,718       & hrs          & 0.0          & 202.0        &       1.5 $\pm$ 2.1               \\
Meal                 & 13,403       & hrs          & 0.0            & 300          &     42.6 $\pm$ 27.0                 \\
Meal Bolus           & 12,863         & hrs        & 0.0          & 18.0         &     4.1 $\pm$ 3.1                 \\
Total Bolus          & 593,089       & $\sim$15 min          & 0.0          & 202.0        &     0.3 $\pm$ 0.9                 \\ \bottomrule
\end{tabular}
\end{table}

\begin{figure}[tb]
  \centering
  \includegraphics[width=0.9\linewidth]{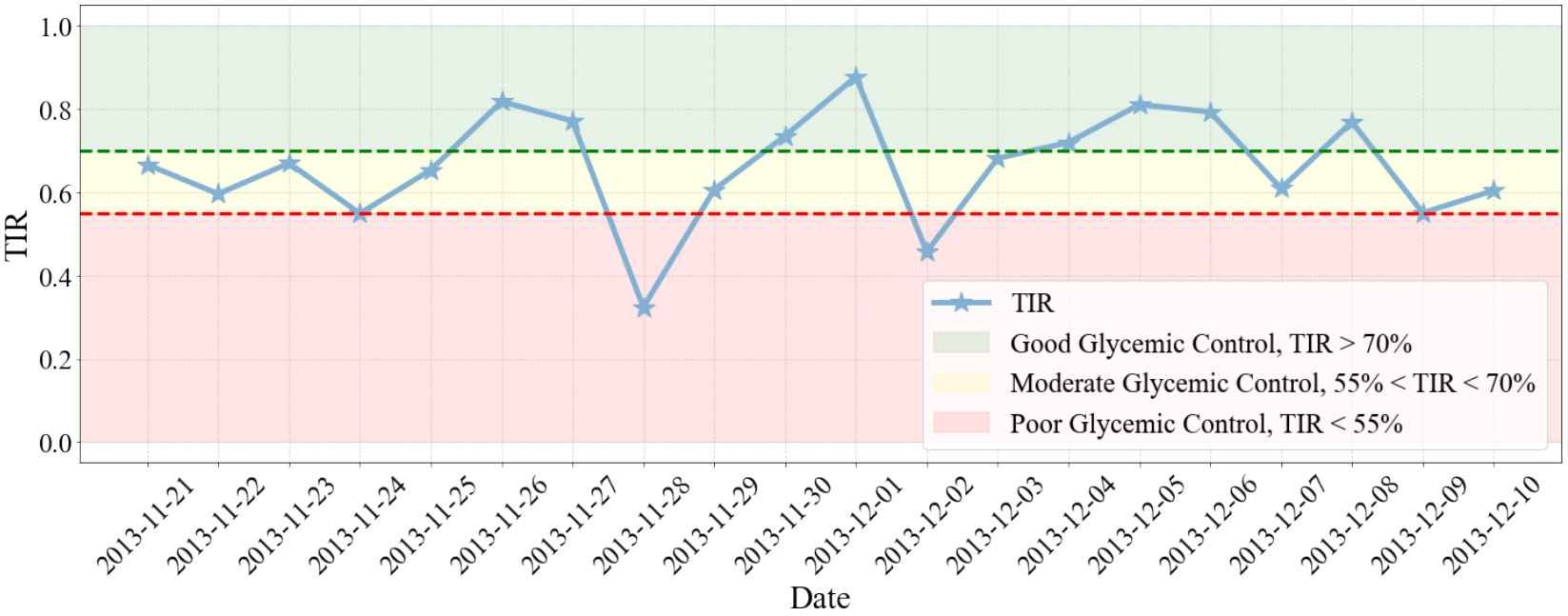}
  \caption{\textbf{
  Visualization of the glycemic control.
  } 
  The glycemic control situation of an example subject is illustrated, containing good, moderate, and poor situations. 
  }
  \label{fig:single_person}
\end{figure}

\begin{table}[htbp]
\centering
\caption{Final Feature Vectors For Model Training.}
\label{table:merged_summary}
\begin{tabular}{@{}ccccc@{}}
\toprule
\textbf{Features} & \textbf{Freq} & \textbf{Min} & \textbf{Max} & \textbf{Mean$\pm$STD} \\ \midrule
TIR                       & 1 day          & 0.00           & 1.00          &      0.74 $\pm$ 0.17                \\
TBR                       & 1 day          & 0.00           & 0.52          &      0.02 $\pm$ 0.04                \\
TAR                       & 1 day          & 0.00           & 1.00          &      0.24 $\pm$ 0.17                \\
Correction Bolus           & 1 day          & 0.00          & 235.70        &    9.36 $\pm$ 8.48                  \\
Meal                     & 1 day          & 1.00            & 926.00          &    170.31 $\pm$ 86.86                  \\
Meal Bolus                  & 1 day        & 0.03          & 56.80         &    15.54 $\pm$ 9.04                  \\
Total Bolus              & 1 day          & 0.10          & 259.70        &     42.40 $\pm$ 19.39                 \\ \bottomrule
\end{tabular}
\end{table}

\subsection{Dataset Processing and Feature Manipulation}
\label{sec:merge}
\subsubsection{Records Merging}
Since the original data in Table \ref{table:summary} are recorded frequently (several minutes per record), they are not suitable for predicting cross-day glucose values. Therefore, we sum all entries within the same day together and merge them into one entry, representing the total amount of data for this patient in a day (e.g. total insulin received in a day). 

\subsubsection{TIR, TBR, and TAR}
Since blood glucose changes frequently and significantly per patient (e.g. blood glucose values decrease rapidly after an injection of insulin), it is unstable and unreliable to directly utilize glucose values (CGM feature) as the model input. 
Therefore, we divide the glucose values into three categories: TIR (time-in-range), TBR (time-below-range), and TAR (time-above-range), following the widely accepted standard criterion \cite{bergenstal2013recommendations}:
\begin{equation}
\label{tir}
    \begin{split}
        &\text{TIR}: 70 \text{ mg/dL} \leq \text{BG} \leq 180 \text{ mg/dL} \\
        &\text{TBR}: \text{BG} < 70 \text{ mg/dL} \\
        &\text{TAR}: \text{BG} > 180 \text{ mg/dL}
    \end{split}
\end{equation}
where BG stands for blood glucose value. The visualization of three ranges of 30 subjects is shown in Figure \ref{fig:cgm_overview}.

After such processing, the frequently changed blood glucose values are converted to the form of intervals, making the model more stable when learning the features.

Therefore, the final feature vectors used for model training are shown in Table \ref{table:merged_summary}, which contains no personal information (e.g. age, race, gender) or physiological data (e.g. blood pressure, heart rate) of the patient.

\subsubsection{Label Generation}
We generate labels for supervising model training and evaluation. Following the standard criterion \cite{bartolome2022computational}, we classify the predicted glucose values into three categories:
\begin{equation}
\label{eqa_label}
    \begin{split}
        &\text{Good Glycemic Control}: \text{TIR} > 0.7 \\
        &\text{Moderate Glycemic Control}: 0.55 \leq \text{TIR} \leq 0.7 \\
        &\text{Poor Glycemic Control}: \text{TIR}  < 0.55
    \end{split}
\end{equation}

Utilizing the three glycemic control categories as clinical guidance can effectively reflect patients' physical condition, providing timely treatment references for them. The glycemic control example of one subject is illustrated in Figure \ref{fig:single_person}.

\begin{table*}[htbp]
\centering
\caption{Comparisons With Several Baselines Under Multiple Evaluation Metrics.}
\label{table:compare}
\resizebox{1.0\linewidth}{!}{
\begin{tabular}{@{}c|ccc|ccc|ccc|cll@{}}
\toprule
\multirow{2}{*}{\textbf{Methods}} & \multicolumn{3}{c|}{\textbf{Good}}                   & \multicolumn{3}{c|}{\textbf{Moderate}}               & \multicolumn{3}{c|}{\textbf{Poor}}                   & \multicolumn{3}{c}{\textbf{Overall}}  \\ \cmidrule(l){2-13} 
                                  & \textbf{Precision} & \textbf{F-1}  & \textbf{Recall} & \textbf{Precision} & \textbf{F-1}  & \textbf{Recall} & \textbf{Precision} & \textbf{F-1}  & \textbf{Recall} & \multicolumn{3}{c}{\textbf{Accuracy}} \\ \midrule
LR                                & 0.71               & 0.81          & 0.94            & 0.39               & 0.25          & 0.18            & 0.31               & 0.08          & 0.05            & \multicolumn{3}{c}{0.67}              \\
RF                                & 0.69               & 0.79          & 0.93            & 0.30               & 0.15          & 0.10            & 0.45               & 0.19          & 0.12            & \multicolumn{3}{c}{0.65}              \\
XGBoost                           & 0.69               & 0.77          & 0.88            & 0.29               & 0.22          & 0.17            & 0.30               & 0.13          & 0.09            & \multicolumn{3}{c}{0.63}              \\ \midrule
\textbf{CrossGP (Ours)}          & \textbf{0.72}      & \textbf{0.82} & \textbf{0.95}   & \textbf{0.48}      & \textbf{0.29} & \textbf{0.21}   & \textbf{0.65}      & \textbf{0.25} & \textbf{0.16}   & \multicolumn{3}{c}{\textbf{0.70}}     \\ \bottomrule
\end{tabular}
}
\end{table*}

\begin{figure*}[tb]
  \centering
  \includegraphics[width=\linewidth]{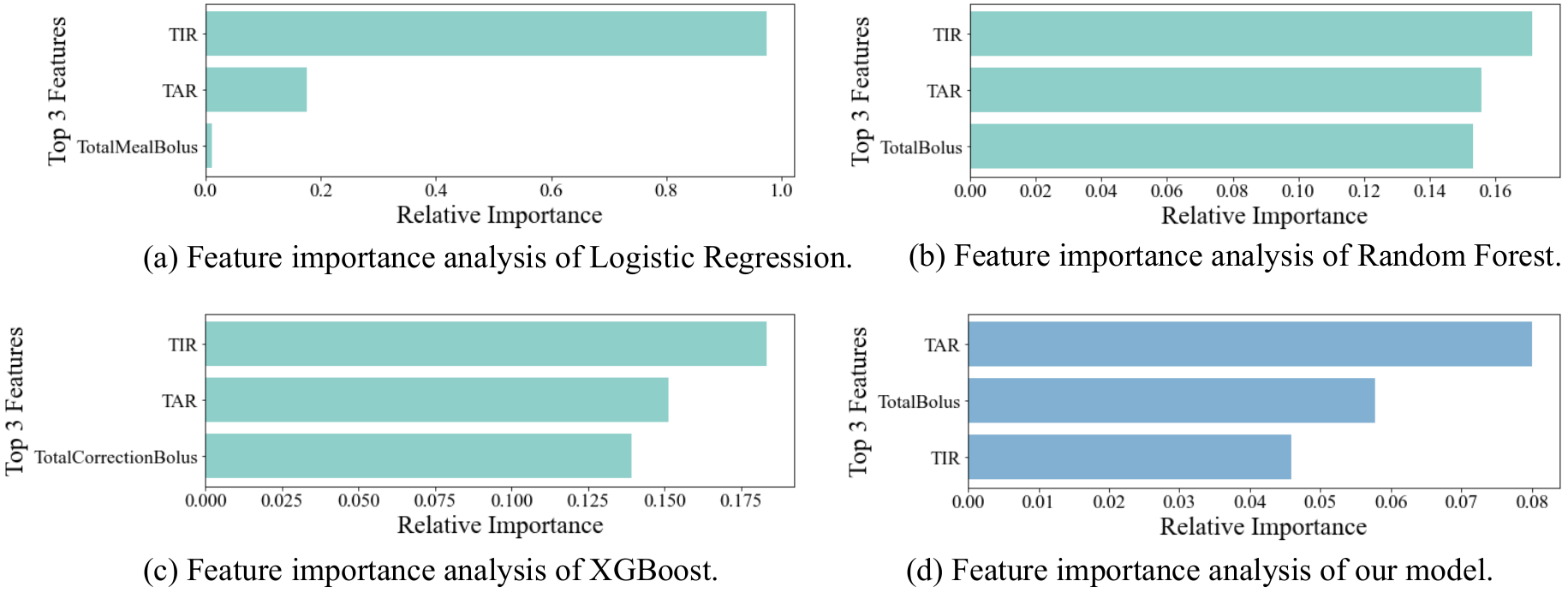}
  \caption{\textbf{
  Visualization of feature importance results.
  } 
  The \textit{top-3} important features learned by each model are demonstrated, where consistent results can be observed, indicating our theoretical assumption of using external activities for glucose prediction is correct. 
  }
  \label{fig:feature}
\end{figure*}

\subsection{Evaluation Metrics}
To evaluate the model's performance, we conduct a classification task for cross-day glucose prediction:
\begin{enumerate}
    \item {
    \textbf{Classification.} 
    Following the criterion in Equation \ref{eqa_label}, the glucose values are classified and labeled into three categories. For each data point of the current day in the test set, the model outputs the predicted category of the next day. Afterward, the classification accuracy, F-1 score, and recall are reported for evaluation. 
    }
    \item {
   \textbf{Feature Importance.}
    To allow potential clinical guidance in practice, we comparatively analyze the weights of features (as their importance) learned by each model when doing prediction.
    Based on the results, we then summarize which features serve as important references for cross-day glucose prediction.
    }
\end{enumerate}

\subsection{Quantitative Results}
The quantitative results are shown in Table~\ref{table:compare}, where our model outperforms existing baselines under all evaluation metrics in each category.
Specifically, for the category ``Good", ``Moderate", and ``Poor", our model reaches the classification precision of $0.72$, $0.48$, and $0.65$, respectively. These accurate predictions demonstrate the potential of the model to provide early warning to diabetic patients in real-life situations without utilizing any personal private data. 
The average precision among all categories ($0.70$) showcases the generalizability and robustness of our model to handle various blood glucose control situations.
As metrics that consider both prediction accuracy and capability of relevant feature capturing, F-1 score ($0.83$) and recall ($0.95$) (e.g. in ``Good" category) reported by our model demonstrate the advantages over all baselines (e.g. $+6.49\%/7.95\%$ of F-1/recall over XGBoost in ``Good" category), indicating the remarkable performance of cross-day glucose prediction.

\subsection{Qualitative Results}
The feature importance of each method is shown in Figure \ref{fig:feature}, where the \textit{top-3} important features learned by the model are illustrated. 
As shown in the figure, consistent results of feature importance analysis are presented by all models: features ``TIR", ``TAR", and ``TotalBolus" occupy the largest weights when making the glucose prediction. 
The qualitative results showcase the possibility of using ``TIR", ``TAR", and ``TotalBolus" as the main components for glucose prediction in the real world, which does not require complex measurement procedures or heavy human resources. Moreover, consistent results given by the four approaches prove the validity of glucose prediction through machine learning technology. 

\section{Conclusion}
In this paper, we propose CrossGP, a novel framework for cross-day glucose prediction without involving the physiological parameters of diabetes patients, providing high-quality predictions while ensuring privacy protection. 
Different from existing work which specializes in either long-term (monthly-based) or short-term (minute-based) predictions, we conduct cross-day predictions solely based on external activities (e.g. insulin dose, meal intake). 
To compare the performance thoroughly, we implement three machine learning baselines (Linear Regression, Random Forest, and XGBoost). Extensive experiments on Anderson's dataset strongly showcase the effectiveness of CrossGP, demonstrating the potential for future clinical applications in the real world. 

%
%
%
\bibliographystyle{splncs04}
\bibliography{mybib}

\end{document}